\documentclass{article}

\usepackage{PRIMEarxiv}

\usepackage[utf8]{inputenc} % allow utf-8 input
\usepackage[T1]{fontenc}    % use 8-bit T1 fonts
\usepackage{hyperref}       % hyperlinks
\usepackage{url}            % simple URL typesetting
\usepackage{booktabs}       % professional-quality tables
\usepackage{amsfonts}       % blackboard math symbols
\usepackage{nicefrac}       % compact symbols for 1/2, etc.
\usepackage{microtype}      % microtypography
\usepackage{lipsum}
\usepackage{fancyhdr}       % header
\usepackage{graphicx}       % graphics
\graphicspath{{media/}}     % organize your images and other figures under media/ folder
\usepackage{tabularx}
\usepackage{booktabs}
\usepackage[title]{appendix}
\usepackage{multirow} 

\usepackage[round,semicolon,authoryear]{natbib}
\usepackage[flushleft]{threeparttable}
\usepackage{babel}

\hypersetup{colorlinks = true, 
linkcolor = black, 
citecolor = black, 
urlcolor = blue}

\title{The Russian Legislative Corpus \\ Russian Law Open Data\thanks{GitHub: \url{https://github.com/irlcode/RusLawOD}; Hugging Face: \url{https://huggingface.co/datasets/irlspbru/RusLawOD}. The authors would like to thank Dmitriy Skougarevskiy, Vladimir Kudryavtsev, and Alex Knorre for their help and encouragement.}}

\author{
  Denis Saveliev, Ruslan Kuchakov \\
  Institute for the Rule of Law \\
  European University at Saint Petersburg \\
  Saint Petersburg\\
  \texttt{\{dsaveliev, rkuchakov\}eu.spb.ru} \\
}

\begin{document}
\maketitle

\begin{abstract}
We present a comprehensive corpus of Russian primary and secondary legislation adopted between 1991 and 2025, comprising 304,382 texts (194,425,905 tokens). The corpus is available in two versions: the basic version contains texts with simple metadata, while the detailed version includes both the original texts and their equivalents converted to the Universal Dependencies CoNLL-U format, annotated with parts of speech, morphological features, and syntactic dependencies. 

\end{abstract}

% keywords can be removed

\section{Introduction}

A comprehensive and up-to-date collection of Russian legislation considered a `gold standard' does not exist, to the best of our knowledge. Here we collect all federal-level primary and secondary legislation covering 1991--2025 and prepare them for downstream natural language processing tasks.

So far limited efforts have been made to create an authoritative and open collection of legal texts in Russian such as monolingual ones (\cite{blinova2022language, blinova2023language}, \cite{athugodage2024transfer}), or parallel corpora \citep{finru_fin, finru_ru}, and as part of a multilingual collection \citep{mulcold}. The limits of these collections are due to several reasons. There are various official sources for publishing legal acts in Russia, and they all have limited consistency. A significant part of legal texts contains only a few words or minor procedural decisions that are technical in nature. In the end, each collection focuses on a specific research question.

Importantly, the prior corpora provide information on the current status of legal acts, such as whether they are still in force, inactive, suspended, or repealed, etc., only as of the date they were downloaded and as provided by the source. Instead, we present all texts as they appeared on the publication date, without any amendments integrated into the original texts.

The remainder of the paper is as follows. In Section 2 we situate the newly built corpus in the existing field of legislation-specific and general legal corpora. In Section 3 we provide a brief institutional context of the Russian legislative procedure and explain how adopted legislation becomes published. In Section 4 we describe our corpus and and descriptive statistics (Section 4) and describe our processing pipeline (Sections 5, 6, 7).

\section{Related corpora}

We present the related legal corpora in  Table~\ref{tab:existing_corpora}. They collect different type of legal texts: statutes (i.e. primary legislation), regulations (i.e. secondary legislation), court decisions, contracts and agreements, transcripts of hearings, meetings, and speeches. Our corpus provides the comprehensive collection of federal legislation including all primary and secondary documents without sampling or tangible exceptions.

\begin{table}[!ht]
	    \caption{Existing corpora of primary and secondary legislation}
	    \label{tab:existing_corpora}
	    \vspace{-0.3cm}
	    \begin{threeparttable}
	    \begin{center}
	    \resizebox{1\textwidth}{!}{\begin{minipage}{\textwidth}
	    \begin{tabular}{>{\raggedright}p{4cm}>{\raggedleft}p{3cm}>{\raggedright}p{3.5cm}>{\raggedright}p{4.5cm}lllcl}
\toprule 
Corpus & Language & Content & Source & \multicolumn{3}{c}{Size} & Period \tabularnewline

 &  & & & Texts & Words/Tokens & GB & \tabularnewline\midrule 

\multicolumn{8}{c}{\emph{Legislation-specific corpora}}\tabularnewline

Corpus of Estonian law texts &  Estonian & statutes & \cite{estonian} & — & 11M(T) & — &  — \tabularnewline

Corpus of Legal Acts of the Republic of Latvia (Likumi) & Latvian & statutes, regulations &\cite{dargis2022}  &  — & 116M(T) & 0.335 & — \tabularnewline

MultiEURLEX & 23 languages  & statutes &\cite{chalkidis2021multieurlex} & 65k & 1.44B(W)& 11.5 & 1958-2016\tabularnewline

Cadlaws & English, French   & statutes &\cite{mauri2021cadlaws}  & — & 35M(T) & — & 2001-2018\tabularnewline

FiRuLex & Finnish, Russian  & — & \cite{finru_fin, finru_ru} & — & 1.5M(T) & — & — \tabularnewline

MULCOLD &  Russian, German, Swedish, Finnish & — & \cite{mulcold} & — & 1M(T) & — & — \tabularnewline

Open Australian Legal Corpus & English & statutes, regulations, judicial decisions & \cite{butler2024} & 229k & 1.4M(T) & 2.35 & — \tabularnewline

Corpus Juridisch Nederlands & Dutch  & statutes & \cite{Odijk2017} & 5.8k  & — & — & 1814-1989\tabularnewline

Rossiyskaya Gazeta Legal Papers  & Russian & statutes & \cite{athugodage2024transfer} & 2k   & — & 0.041 & 2008-2022\tabularnewline

RusLawOD  & Russian & statutes, regulations, judicial decisions  & Ours \citep{saveliev2018} & 281k & 176M(T) & 21 & 1991-2023\tabularnewline \hline

\multicolumn{8}{c}{\emph{General-purpose legal corpora that include legislation}}\tabularnewline

Corpus Juris Rossicum Contemporaneum & Russian  & statutes, regulations, judicial decisions, transcripts, papers, blogs  & \cite{blinova2022language, blinova2023language} & 44k & 119M(W) & 118 & — \tabularnewline

MultiLegalPile  & 24 languages & statutes, regulations, judicial decisions, contracts, transcripts &\cite{niklaus2023multilegalpile} & — & 87B(W) & 689 & — \tabularnewline

Pile of Law & mainly English & statutes, regulations, judicial decisions, contracts, transcripts & \cite{hendersonkrass2022pileoflaw} & 10M & 30B(W) & 256 & — \tabularnewline

Legal Documents from Norwegian Nynorsk Municipialities & Norwegian & municipal acts, transcripts & \cite{kasen2020} & 50k & 127M(W)  & 0.9 & 2010-2020\tabularnewline
\bottomrule
\end{tabular}
     \emph{\footnotesize{}Note:}{ In brackets `W' — words, `T' — tokens.}
	    \end{minipage}}
     \end{center} 
        \end{threeparttable}
\end{table}

%\begin{itemize}
    %\item The Corpus of `Local' Acts (\texttt{CorRIDA}): This collection contains 1,546 documents totaling 1.784 million tokens and legal documents that typically concern laypeople.
    %\item The Corpus of Constitutional Court Decisions (\texttt{CorDec}): This collection comprises 584 documents with 3,427 tokens. 
    %\item The Corpus of Legal Acts from significant Federal Laws, Codes, and Government Decrees (\texttt{CorCodex}): This collection includes 279 documents amounting to 3.227 million tokens.
%\end{itemize}

% These three corpora and extensions are compiled together in one project titled \href{https://plaindocument.spbu.ru}{`Corpus Juris Rossicum Contemporaneum.'} (\texttt{CoJurRC}). Our work differs in two distinct ways. First, these corpora focus on specific samples of legal documents. In contrast, our work covers the entire universe of federal legal acts. Second, we compile a rich set of metadata.   

\section{Historical context of publication routine and official sources of legal information}

Russia is a civil law country. This means that its main source of law comes from the elected legislative body (the Federal Assembly), in the form of laws and codes, which are the primary source of legislation. The President, as an independent authority, and the executive branch led by the government, may adopt legal norms or individual-oriented acts, such as presidential decrees, governmental and ministerial rulings, and other acts of the executive branch, which become secondary legislation. Precedents and court decisions are not generally considered a source of normative legislation, although the decisions of the highest courts can influence the practice of lower courts to ensure uniformity, and decisions of the Constitutional Court can rule on the constitutionality of laws. Subjects of the Russian Federation have their own legislation on certain topics mentioned in the Constitution of Russia. Therefore, the present corpus consists of federal Russian laws, codes, and the Constitutional Court's acts as primary law, and other decrees, rulings, and provisions as secondary law.

Since the USSR disintegration in 1991, Russian legislation were almost completely rewritten, so that we don't need to include legislation adopted before that year in our corpus for it to correctly represent the legislation currently in force.  

\begin{figure}[h]
  \centering
  \caption{Number of Legal Acts Adopted Annually, 1991–2025}
  \includegraphics[width=\textwidth]{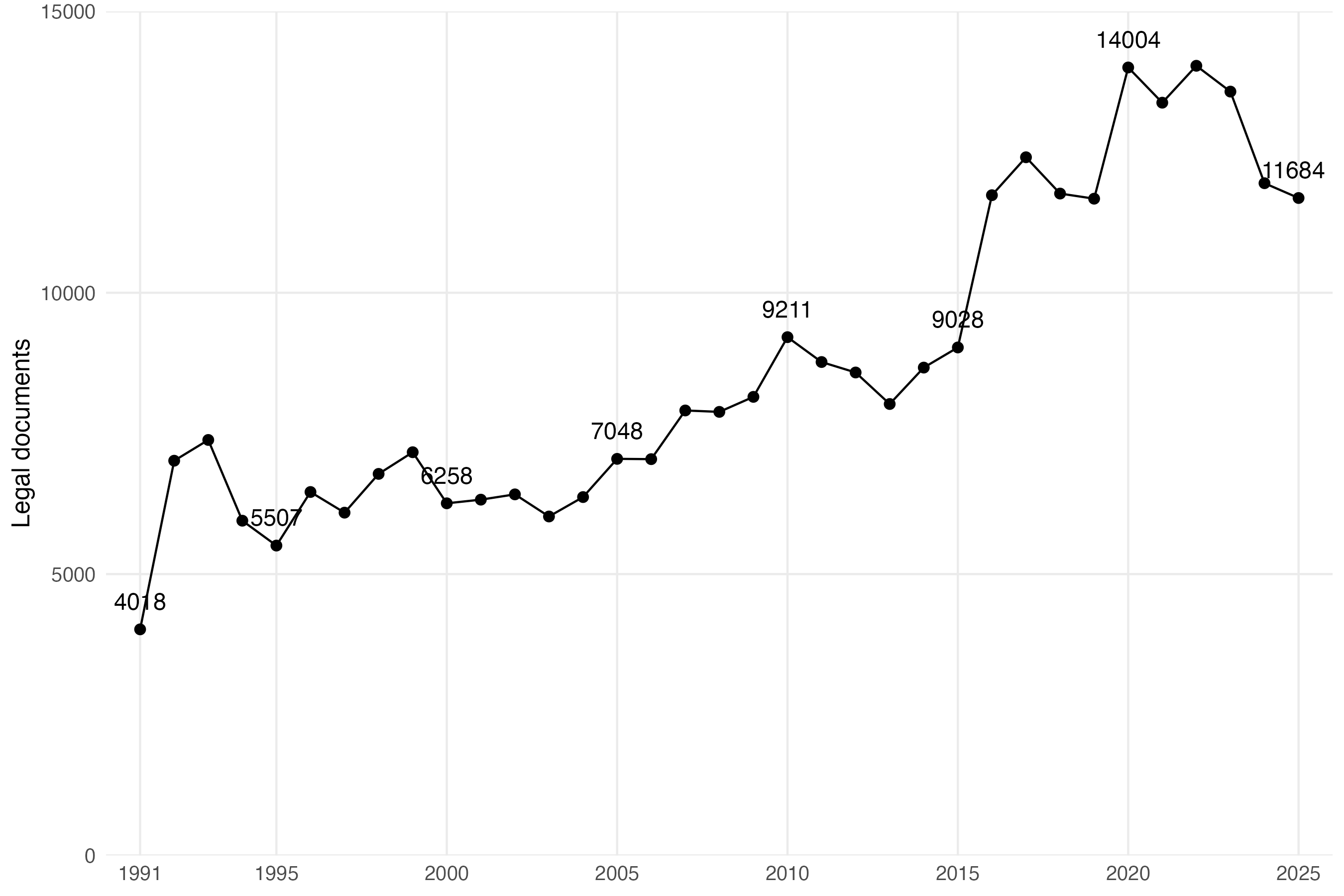}
  \label{fig:fig1}
  \vspace{-20pt}
\end{figure}

\paragraph{Public sources of legislation}

Russian legislation of the Soviet era was officially published in paper bulletins. But the Soviet authorities did not publish most secondary legislation, despite it being legally binding. This circumstance gave some room for arbitrariness. In 1990, the Committee for Constitutional Supervision of the USSR established\footnote{\texttt{https://base.garant.ru/1777169/}} that not officially published for general information were not applicable. The obligatory publication of normative acts became one of the most important guarantees of human and citizens' rights. Finally the obligatory official publication of the whole universe of legislation (excluding only those made State Secret) opened the possibility to have the open corpus of legal acts in the electronic form. 

Following the obligatory publication of legal normative acts, publications of modern Russian legislation first appeared only in paper format in the official gazettes `Rossiyskaya Gazeta'\footnote{\texttt{https://rg.ru/doc}} and `Vedomosti Verhovnogo Soveta RSFSR.'\footnote{\texttt{http://www.szrf.pravo.gov.ru/}} In the 1980s, the Ministry of Justice of the USSR started developing the `standard electronic bank of legislation.' The plan aimed to compile every legal act, with each act's text serving as the `golden standard.' However, this plan failed.

It would be an understatement to call the official publication process in modern Russia scattered. First, paper versions of legislation are still published in the Russian gazette `Rossiyskaya Gazeta.' Second, the subordinated agency of the Federal Protective Service of Russia called the Special Communications Service (`Spetssvyaz FSO') handles the publication of both paper and electronic versions of the official bulletin `The Collection of Legislation of Russian Federation'. Third, this federal agency also maintains an electronic system with reference versions of legislation called `The Integrated Bank `Legislation of Russia'' as a part of its web site \texttt{pravo.gov.ru}. Fourth,  since 2011, the agency has been operating a separate online official electronic publication system called `The Official Publication of Legal Acts'. Fifth, the Ministry of Justice of Russia provides access to legal documents through its online service `Normative acts of Russian Federation' at \texttt{pravo.minjust.ru}. Sixth, most public authorities required to publish legislation on their websites, with some also available in paper form.

Here we need to be more clear on what is `official' in terms of Russian laws and what isn't. Official publication mean the authoritative text of a legal act exactly as it was signed, by which anyone, especially officials (judges, etc), may confirm the act authenticity. Mostly any act comes into force after a certain number of days after the first official publication. The date of first official publication of any act may be of high importance to law enforcement. The official source is the source directly mentioned in special laws as such. Any other sources may publish laws as referential.

%This cast some doubts. How to check if really the appropriate state official signed manually the exact text? Any legal document need to be signed by an appropriate person in the name of the state body. At first it was always a paper document printed with a typewriter (later — computer printer) and signed manually in person. But for it to widely spread it has to be published in many copies and sent all over the country. Therefore, first, the office of the state organ is responsible to print the necessary amount of copies to send them to other state organs, and the official gazette (or other sources of the official publication). Second, the official gazette, bulletin or other source makes more copies for libraries and other readers. Copies can't be exact in pixels, they may be of other font type, layout, formatting etc. And they do not include handwritten signature of the signatory. As the result, only a state archive have the real signed document, others see something else which authority need to be confirmed. At least, in the paper era, many copies in different libraries were the guarantee that the text wasn't changed.

%The things become much more doubtful with the electronic publication. Official internet portal \texttt{pravo.gov.ru} does not publish neither e-signature of the signatory person, nor its own e-signature. We can suggest that electronic text on that portal may arbitrarily change without notice after the publication (but we are not claiming that they do). Another drawback of the official publication portal is 

As the result of our research, we choose the `The Integrated Bank `Legislation of Russia''  as a source of texts and metadata for our corpus by two reasons. First, though it is not official publication in terms described above, it is published by a state organ and represent acts for the maximal covering period. Second, despite it doesn't have an API, it can be queried by http requests for data and documents.

\paragraph{Private sources of legislation}

Legal professionals and laypeople generally rely on private legal information systems. `Consultant Plus'\footnote{\texttt{https://www.consultant.ru}} and `Garant'\footnote{\texttt{https://www.garant.ru}} are among the most popular ones. These systems contain hundreds of millions of legal texts including primary and secondary legislation with all the the amendments, thus providing access to historical versions of those acts. The underlying data, despite their superior quality, are proprietary and unavailable to the general public in the form of a free downloadable corpus.

\paragraph{Amendments and versions of the legal acts, legal force}

Researchers working with the legal text corpus must take into account the legal status of the acts and amendments. When state authorities amend a legal act, they do not necessarily officially adopt and publish an entirely new text. More often, amendments are made to existing texts, specifying which parts to add, delete, or change. To create a consolidated version of a legal act that has legal force on a specific date, researchers must apply the relevant amendments to the original text. Some legal acts lose their legal force due to the expiration of the period specified by the legislator or because they are repealed.

Including consolidated versions of legal documents incorporating all amendments would be labor-intensive, lead to text duplication, and distort metrics such as token counts. Providing users with up-to-date information on the legal status of each act is also continuously labor-intensive. Thus, the corpus cannot serve as an up-to-date source of current law. However, we provide all available amendment texts, enabling researchers to track changes if needed and observe the changing discourse.

\section{Corpus building}

We queried the chosen source -- `Legislation of Russia' web site for several periods spanning from 2017 to 2026 with the last data collection taking part in February 2026.

\paragraph{Sentence segmentation}

Legal text is noticeably different from general domain text: fewer exclamatory or interrogative sentences, more abbreviations with various punctuation marks. Applying a general-purpose sentence segmentation procedure on legal texts is known to yield suboptimal results. Instead, we relied on rule-based approach to sentence segmentation. We kept only complete sentences in the text. First, we removed dots that constituted common abbreviations using regular expressions. Then, we split the text, considering a fragment ending with a dot or semicolon to be a sentence. Question marks and exclamation marks do not end sentences in legal documents. The semicolons in this type of text usually indicate a list item.

\paragraph{Removing uninformative blocks}

Legislation text has several mandatory blocks that are effectively metadata, such as headers with certain details (date, name of state body), blocks designating appendixes and the final block with signatures. These blocks may take up more space for some smaller documents than the actual text. We removed all such uninformative blocks with regular expressions and a set of rules.

\paragraph{Removing arrays of digits}

The legal corpus collects budget documents and lists of border coordinates (e.g. borderlines between regions and municipalities). These documents may count hundreds of pages full of bare digits. We discarded all such documents. In the documents with texts containing long numbers for morpho-syntactic markup we replaced any number with `999' to represent it properly as a numeral.

\paragraph{Unifying formatting}

The legal text may contain tables. Another problem is that the data source stores legal documents in inconsistent formats that have changed over time. For example, legislation adopted before 2017 had HTML-based layout elements, inconsistent indentation, and pseudographic tables. After 2017, HTML coding began to be used more consistently, with tables compatible with standard HTML layouts. We removed all tables to make the texts more homogeneous. We also used regular expressions to remove extra punctuation, indents, spaces, and technical references.

\paragraph{Flattening lists}

One marked feature of legal texts is it list-based structure. We flattened each list and represent each item as a separate sentence.

Figure \ref{fig:fig1} compares the size of the original corpus size and its processed version. We observe 9,991,958,759 symbols in the original version which are reduced to 1,544,031,903 symbols in the processed version .

\paragraph{Core legal texts}

The public significance of different legal documents varies widely, as does their scope. We identified the most significant acts of the legal corpus constituting its core. The core excludes all procedural documents, or legal acts that only amend other laws. We identified such documents with a rule-based procedure. First, we selected only those documents that we considered to be normative, such as codes, laws, presidential decrees, and government regulations. Then, we excluded any documents that amended other documents or made them out of force. Finally, we also excluded documents that contained specific words in their titles that clearly indicated they were not normative documents, like awards for state prizes. Figure \ref{fig:fig2} shows the annual dynamics of adopted acts and their tokens for the entire corpus and a sample of the core acts.

\begin{figure}[ht]
  \centering
  \caption{Yearly corpus size, symbols, before/after preprocessing, all legislation and core legislation}
  \includegraphics[width=\textwidth]{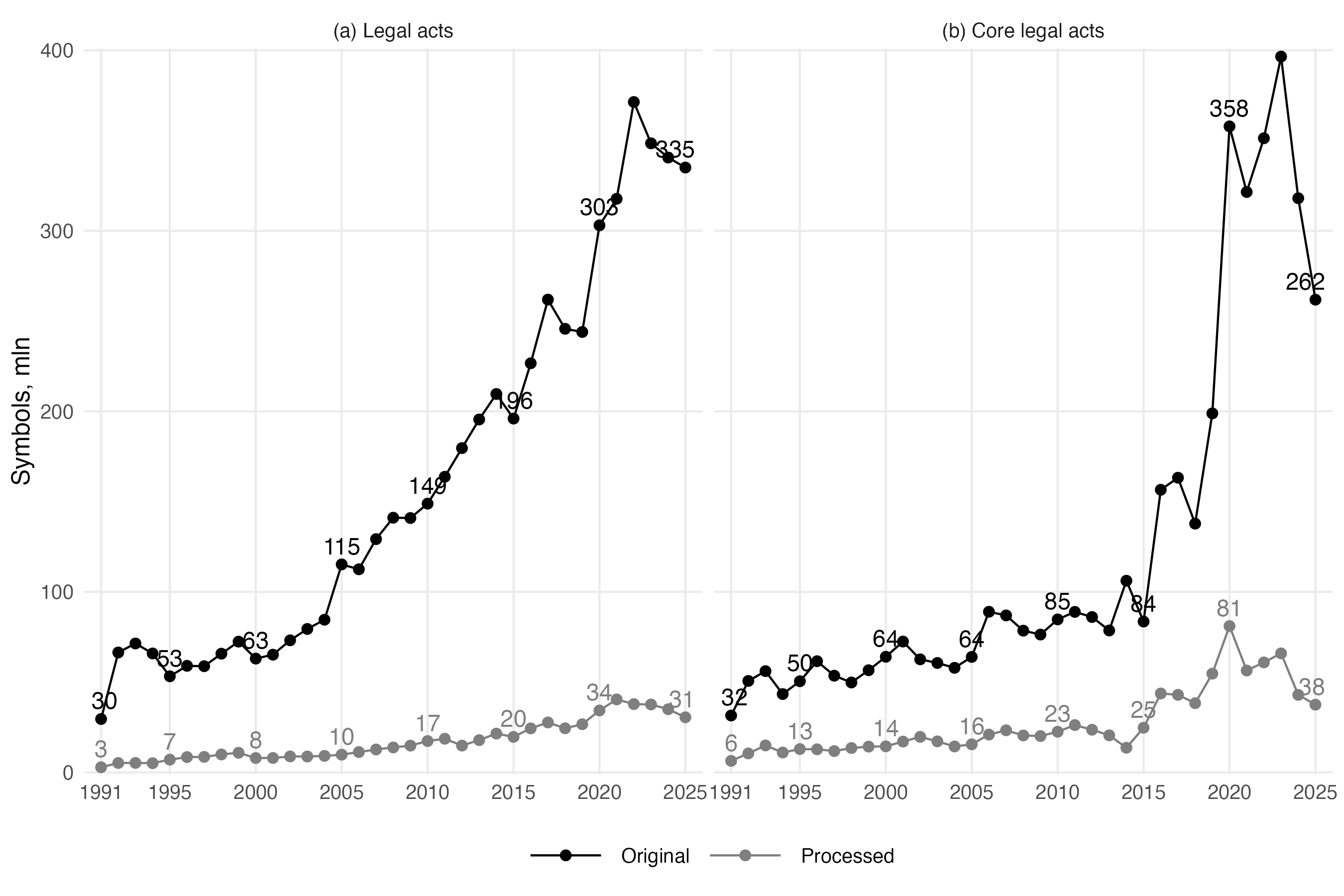}
  \label{fig:fig2}
  \vspace{-20pt}
\end{figure}

%Figure \ref{fig:fig1} shows the dynamics of adopted legal acts and their tokens. The average number of normative acts adopted yearly is 4.9\% higher than the previous year. The volume of these acts increases by 9.8\% annually.

%\begin{figure}[!ht]
%  \centering
%  \caption{The number of adopted documents and its tokens per year.}
%  \includegraphics[width=16cm]{figures/plot_doc_tokens_annually.png}
%  \label{fig:fig1}
%\end{figure}

\begin{figure}[h]
  \centering
  \caption{Example of XML structure}
  \includegraphics[width=14cm]{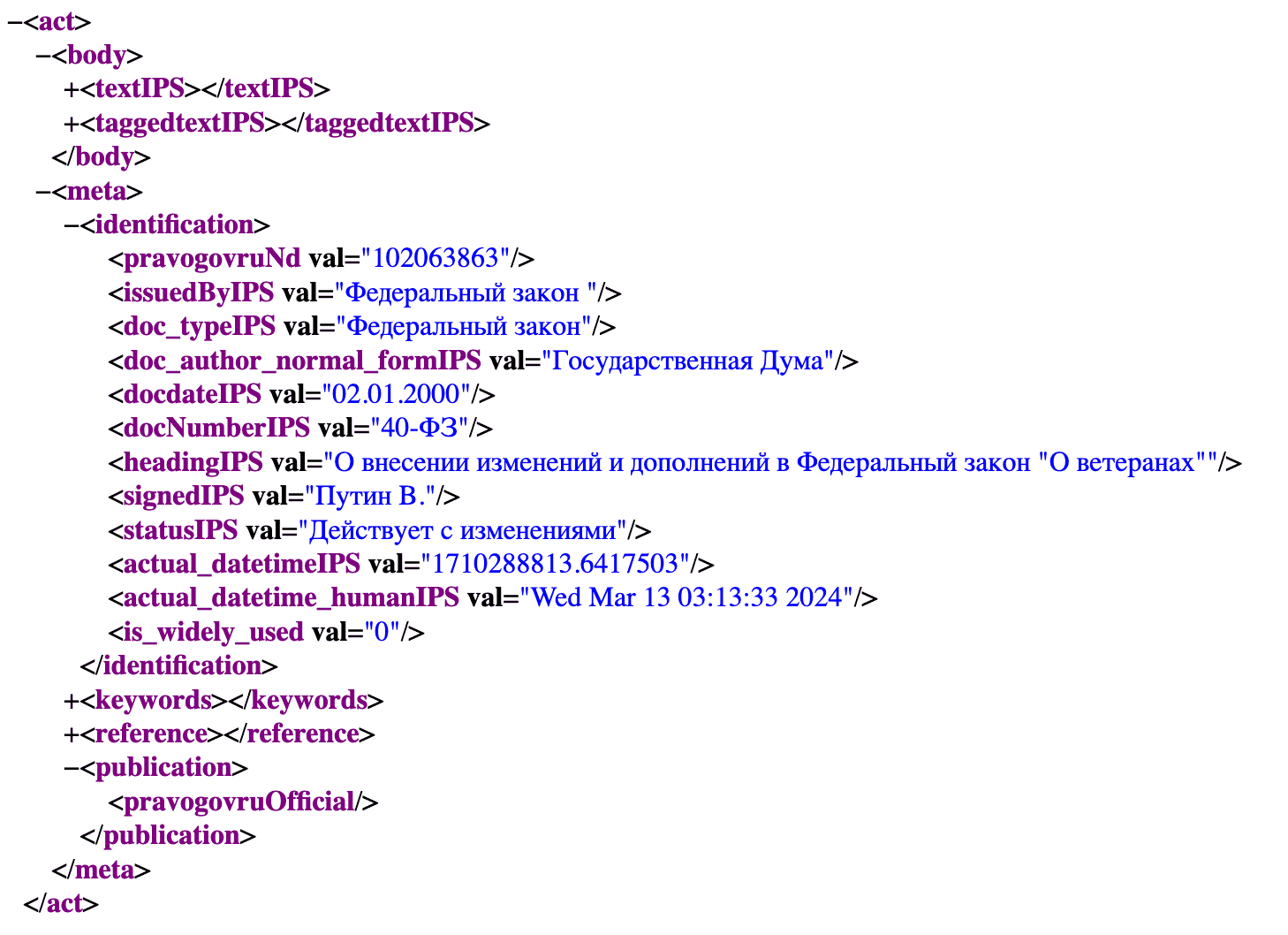}
  \label{fig:fig2}
\end{figure}

\paragraph{Processing}

Each final piece of legislation is saved as an XML file. We follow the Akoma Ntoso standard / OASIS LegalDocML \citep{akoma2018} for the structure and naming. The tags <body> and <meta> contain the text itself and metadata attributes, respectively. The structure of a typical XML document is presented in Figure \ref{fig:fig2}.  

For lemmatization, POS tagging, and dependency parsing we use the \texttt{ru-syntax}\footnote{\texttt{https://github.com/tiefling-cat/ru-syntax}} tool developed by the Computational Linguistics team at the Higher School of Economics. This tool gathers the results from the morphological analyzer \texttt{MyStem}\footnote{\texttt{https://yandex.ru/dev/mystem/}}, \texttt{part-of-speech tagging} \citep{schmid2013probabilistic} \texttt{TreeTagger}\footnote{\texttt{https://www.cis.lmu.de/~schmid/tools/TreeTagger/}}, and dependency grammar analyzer \texttt{MaltParser}\footnote{\texttt{http://www.maltparser.org}}. Finally, the result is stored in the CONLL-U format and saved as XML.

Table \ref{tab:metadata} contains the list of available metadata. We did not change `keywords' and `references' created by officials. A reference is an official classification\footnote{\texttt{http://pravo.gov.ru/proxy/ips/?docbody=\&prevDoc=102027655\&backlink=1\&\&nd=102064921}} of topics related to legal acts with a limited number of choices. Each document has multiple `keywords' and `reference'. 

\begin{table}[!ht]
	    \caption{Available Metadata}
	    \label{tab:metadata}
	    \vspace{-0.3cm}
	    \begin{threeparttable}
	    \begin{center}
	    \resizebox{1\textwidth}{!}{\begin{minipage}{\textwidth}
	    \begin{tabular}{lllrrr}
\multicolumn{1}{c}{Node} & \multicolumn{1}{c}{Attribute} & \multicolumn{1}{c}{Comment}                                                                        & \multicolumn{1}{c}{N} & \multicolumn{1}{c}{Unique, n} & \multicolumn{1}{c}{Missing, \%} \\\toprule\toprule
body                     & textIPS                       & the text of a legal act                                                                            & 304857                & –                             & 0.9\footnote{We ignored texts counting only a few words or a sentence piece.}                            \\
                         & taggedtextips                 & \begin{tabular}[c]{@{}l@{}}Tagged text, CONLL-U\end{tabular}                 & 304857                & –                             & 0.9                            \\
meta                     & pravogovruNd                  & \begin{tabular}[c]{@{}l@{}}Internal ID, pravo.gov.ru \end{tabular}          & 304857                & 304857                        & 0     \\

                         & issuedByIPS                   & \begin{tabular}[c]{@{}l@{}}Full name of a legal act \\ (type + state body)\end{tabular}     & 304857                & 2104                          & 0                              \\
                         
                         & doc\_typeIPS                  & Type of legal act                                                                              & 304857                & 67                            & 0                              \\
                         
                         & doc\_author\_normal\_formIPS  & \begin{tabular}[c]{@{}l@{}}Unified state body name\end{tabular}          & 304857                & 296                           & 0.2                            \\
                         & docdateIPS                    & Date of official publication                                                                       & 304857                & 11228                         & 0                              \\
                         
                         & docNumberIPS                  & Document number at signature                                                                       & 304857                & 80267                         & 0                              \\
                         & headingIPS                    & Title                                                                                              & 304857                & 243664                        & 0                              \\
                         & signedIPS                     & Official  who signed an act                                                                       & 304857                & 1360                          & 11.2                            \\
                         
                         & statusIPS                     & Status                                                                                             & 304857                & 4                             & 0                              \\
                         
                         & actual\_datetimeIPS           & Dump date                                                                                 & 304857                & 287297                        & 0.4                            \\
                         
                         & actual\_datetime\_humanIPS    & Dump date                                                                                  & 304857                & 123215                        & 0.4                            \\
                         
                         & is\_widely\_used              & \begin{tabular}[c]{@{}l@{}} Flag for core acts\end{tabular} & 304857                & 2                             & 0                              \\
keywords                 & keywordsByIPS                 & key words                                                                                          & 304857                & 175495                        & 19.7                           \\
reference                & classifierByIPS               & official classifier                                                                       & 304857                & 134404                        & 20.0 \\\bottomrule\bottomrule                         
\end{tabular}

	    \end{minipage}}
     \end{center} 
        \end{threeparttable}
\end{table}

%\begin{figure}[!ht]
%  \centering
%  \caption{The number of adopted documents and their tokens per year by samples.}
%  \includegraphics[width=16cm]{figures/plot_doc_tokens_annuallyb.png}
%  \label{fig:fig2}
%\end{figure}

%\section{Applications}

%\subsection{Readability assessment}

%One natural use of the created corpus is the readability assessment task \cite{kuchakov2018, saveliev2020}.

\section{Access and usage}

Russian law excludes texts of legal acts from copyright protection, so they can be distributed freely. Official publication metadata are subject to terms at the source site, allowing both commercial and non-commercial use, providing attribution to the source. Other project materials are distributed under the Creative Commons Attribution-NonCommercial 4.0 International license. The project repository is \url{https://github.com/irlcode/RusLawOD}. The dataset is also available at Hugging Face: \url{https://huggingface.co/datasets/irlspbru/RusLawOD}. 

\bibliographystyle{chicago}
\bibliography{references}  

@article{blinova2023language,
	title = {Language complexity across sub-styles and genres in legal {R}ussian},
	volume = {9},
	issn = {2313-8912},
	doi = {10.18413/2313-8912-2023-9-2-0-5},
	number = {2},
	journal = {Research Result. Theoretical and Applied Linguistics},
	author = {Blinova, Olga V and Tarasov, Nikita A},
	year = {2023},
	pages = {73--96},
    note = {10.18413/2313-8912-2023-9-2-0-5}
}

@article{blinova2022language,
    title = {A hybrid model of complexity estimation: {Evidence} from {Russian} legal texts},
    volume = {5},
    issn = {2624-8212},
    url = {https://www.frontiersin.org/articles/10.3389/frai.2022.1008530/full},
    note = {10.3389/frai.2022.1008530},
    journal = {Frontiers in Artificial Intelligence},
    author = {Blinova, Olga and Tarasov, Nikita},
    year = {2022},
}

@article{saveliev2018,
  title={On Creating and Using Text of the {Russian} {Federation} Corpus of Legal Acts Acts as Open Dataset},
  author={Saveliev, Denis},
  journal={Pravo. Zhurnal Vysshey shkoly ekonomiki},
  number={1},
  pages={26--44},
  year={2018},
  note={10.17323/2072-8166.2018.1.26.44 (in Russian)}
}

@article{chalkidis2021multieurlex,
  title={MultiEURLEX--A multi-lingual and multi-label legal document classification dataset for zero-shot cross-lingual transfer},
  author={Chalkidis, Ilias and Fergadiotis, Manos and Androutsopoulos, Ion},
  journal={arXiv preprint arXiv:2109.00904},
  year={2021}
}

@inproceedings{athugodage2024transfer,
  title={Transfer Learning for {R}ussian Legal Text Simplification},
  author={Athugodage, Mark and Mitrofanove, Olga and Gudkov, Vadim},
  booktitle={Proceedings of the 3rd Workshop on Tools and Resources for People with REAding DIfficulties (READI)@ LREC-COLING 2024},
  pages={59--69},
  year={2024}
}

@inproceedings{akoma2018,
  title={Akoma Ntoso Version 1.0 Part 1: XML Vocabulary},
  author={Palmirani, Monica and Sperberg, Roger and Vergottini, Grant and Vitali, Fabio},
  year={2018},
  note={\url{http://docs.oasis-open.org/legaldocml/akn-core/v1.0/akn-core-v1.0-part1-vocabulary.html}}
}

@inproceedings{schmid2013probabilistic,
  title={Probabilistic part-of-speech tagging using decision trees},
  author={Schmid, Helmut},
  booktitle={New methods in language processing},
  pages={154--164},
  year={2013},
  organization={Routledge}
}

@misc{hendersonkrass2022pileoflaw,
  url = {https://arxiv.org/abs/2207.00220},
  author = {Henderson, Peter and Krass, Mark S. and Zheng, Lucia and Guha, Neel and Manning, Christopher D. and Jurafsky, Dan and Ho, Daniel E.},
  title = {Pile of Law: Learning Responsible Data Filtering from the Law and a 256GB Open-Source Legal Dataset},
  publisher = {arXiv},
  year = {2022}
}

@article{niklaus2023multilegalpile,
  title={Multilegalpile: {A} 689gb multilingual legal corpus},
  author={Niklaus, Joel and Matoshi, Veton and St{\"u}rmer, Matthias and Chalkidis, Ilias and Ho, Daniel E},
  journal={arXiv preprint arXiv:2306.02069},
  year={2023}
}

@article{mauri2021cadlaws,
  title={Cadlaws--An {E}nglish--{F}rench Parallel Corpus of Legally Equivalent Documents},
  author={Mauri, Francina Sole and S{\'a}nchez-Gij{\'o}n, Pilar and Gonz{\'a}lez, Antoni Oliver},
  journal={Mutatis Mutandis: Revista Latinoamericana de Traducci{\'o}n},
  volume={14},
  number={2},
  pages={494--508},
  year={2021},
  publisher={Universidad de Antioquia}
}

@book{Odijk2017,
 address = {London},
 doi = {10.5334/bbi},
 editor = {Odijk, Jan and van Hessen, Arjan},
 isbn = {978-1-911529-24-8, 978-1-911529-25-5, 978-1-911529-26-2, 978-1-911529-27-9},
 month = {Dec},
 pages = {414},
 publisher = {Ubiquity Press},
 title = {CLARIN in the Low Countries},
 year = {2017}
}

@article{estonian,
  author={Kaalep, Heiki-Jaan},
  title={Corpus of {E}stonian law texts},
  journal={Estonian Legal Language Centre},
  note={\url{https://www.cl.ut.ee/korpused/segakorpus/seadused/}},
  year={2002}
}

@misc{dargis2022,
 title = {Corpus of Legal Acts of the {R}epublic of {L}atvia ({L}ikumi)},
 author = {Darģis, Roberts},
 publisher = {{CLARIN}-{LV} digital library at {IMCS}, University of {L}atvia},
 copyright = {Creative Commons - Attribution 4.0 International ({CC} {BY} 4.0)},
 year = {2022},
 note={\url{http://hdl.handle.net/20.500.12574/65}}}

@misc{finru_fin,
 author={{University of Tampere}},
 year={2011},
 title={{The Finnish Sub-corpus of FiRuLex, {R}ussian-{F}innish Comparable Corpus of Legal Texts}},
 publisher={Kielipankki},
 type={data set},
 note={\url{https://www.kielipankki.fi/korp/#?corpus=legal_fi&cqp=%5B%5D}},
}

@misc{finru_ru,
 author={{University of Tampere}},
 year={2011},
 title={{The {R}ussian Sub-corpus of FiRuLex, {R}ussian-{F}innish Comparable Corpus of Legal Texts}},
 publisher={Kielipankki},
 type={data set},
 note={\url{https://www.kielipankki.fi/korp/?mode=other_languages#?corpus=legal_ru&prequery_within=sentence&cqp=%5B%5D}},
}

@article{mulcold,
  author={{University of Tampere}},
  title={MULCOLD — Multilingual Corpus of Legal Documents},
  year={2013},
  note={\url{https://www.kielipankki.fi/korp/?mode=parallel#?parallel_corpora=fin&corpus=mulcold_fi&cqp_fi=%5B%5D&cqp_fin=%5B%5D}}
}

@misc{butler2024,
    author = {Butler, Umar},
    year = {2024},
    title = {Open {A}ustralian Legal Corpus},
    publisher = {Hugging Face},
    version = {7.0.4},
    doi = {10.57967/hf/2833},
    url = {https://huggingface.co/datasets/umarbutler/open-australian-legal-corpus}
}

@misc{kasen2020,
    author ={Kåsen, Andre},
    title={Legal Documents from {N}orwegian {N}ynorsk {M}unicipialities},
    year = {2020},
    title = {Open Australian Legal Corpus},
    publisher = {National Library of Norway},
    note = {\url{https://www.nb.no/sprakbanken/en/resource-catalogue/oai-nb-no-sbr-60/}}
}

\clearpage

\end{document}